\documentclass{article} % For LaTeX2e
\usepackage[final]{colm2026_conference}
\usepackage{multirow}
\usepackage{microtype}
\usepackage{url}
\usepackage{booktabs}
\usepackage{array}
\usepackage{subcaption}
\usepackage{colortbl}       % for \rowcolor
\usepackage{xspace}
\usepackage[table]{xcolor}  % for gray!15 color syntax

\usepackage{xcolor}

\usepackage{colortbl}
\usepackage{enumitem}
\usepackage{amsmath,amssymb}
\usepackage{caption}
\usepackage{wrapfig}
\usepackage{pifont}
\newcommand{\cmark}{\ding{51}}
\newcommand{\xmark}{\ding{55}}

\usepackage[colorlinks,citecolor=blue,urlcolor=blue,linkcolor=blue]{hyperref}

\usepackage{lineno}

\usepackage{cleveref}

\usepackage{graphicx}

\newcommand{\benchmark}{\textsc{EgoMemReason}}

\newcommand{\worldwideweb}{\raisebox{-1.5pt}{\includegraphics[height=1.05em]{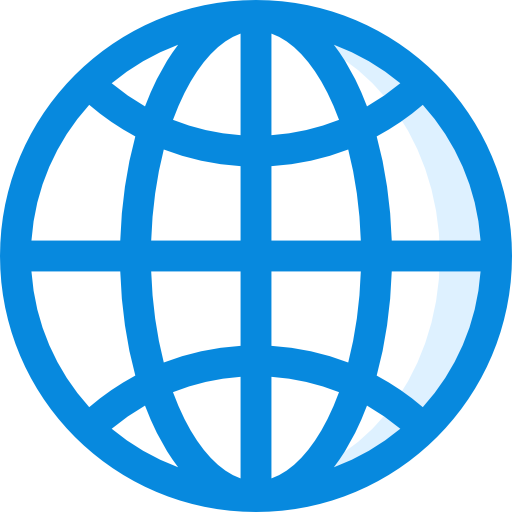}}\xspace}
\newcommand{\github}{\raisebox{-1.5pt}{\includegraphics[height=1.05em]{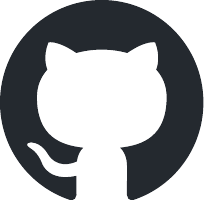}}\xspace}

\definecolor{darkblue}{rgb}{0, 0, 0.5}
\hypersetup{colorlinks=true, citecolor=darkblue, linkcolor=darkblue, urlcolor=darkblue}

\title{
\benchmark{}: A  Memory-Driven Reasoning Benchmark
for Long-Horizon Egocentric Video Understanding
}

\author{%
\textbf{Ziyang Wang}$^{1,}$\thanks{Equal contribution.} \quad 
\textbf{Yue Zhang}$^{1,*}$ \quad
\textbf{Shoubin Yu}$^1$ \quad
\textbf{Ce Zhang}$^1$ \quad
\textbf{Zengqi Zhao}$^1$ \\
\textbf{Jaehong Yoon}$^{2}$ \quad
\textbf{Hyunji Lee}$^1$ \quad
\textbf{Gedas Bertasius}$^1$ \quad
\textbf{Mohit Bansal}$^1$ \\
\\$^1$UNC Chapel Hill \quad\quad $^2$NTU Singapore\\
\\
}

\begin{document}

\ifcolmsubmission
\linenumbers
\fi

\maketitle

\vspace{-10mm}
\begin{flushleft}
\github \href{https://github.com/Ziyang412/EgoMemReason}{\text{Code}} \quad\quad
\worldwideweb \href{https://egomemreason.github.io/}{\text{Project page}} \quad\quad
\end{flushleft}

\begin{abstract}
Next-generation visual assistants, such as smart glasses, embodied agents, and always-on life-logging systems, must reason over an entire day or more of continuous visual experience.
In ultra-long videos, relevant information is sparsely distributed across hours or days, making \textbf{memory a fundamental challenge}: models must accumulate information over time, recall prior states, track temporal order, and abstract recurring patterns.
However, existing week-long video benchmarks are primarily designed for perception and recognition, such as moment localization or global summarization, rather than reasoning that requires integrating evidence across multiple days.
To address this gap, we introduce \textbf{\benchmark{}}, a comprehensive benchmark for week-long egocentric video understanding through memory-driven reasoning.
\benchmark{} evaluates three complementary memory types: \textbf{entity memory}, tracking how object states evolve and change across days; \textbf{event memory}, recalling and ordering activities separated by hours or days; and \textbf{behavior memory}, abstracting recurring patterns from sparse, repeated observations over the whole week period. 
\benchmark{} comprises 500 questions across three memory types and six core challenges, with an average of 5.1 video segments of evidence per question and 25.9 hours of memory backtracking.
We evaluate \benchmark{} on \textbf{17} methods across MLLMs and agentic frameworks, revealing that even the best model achieves only 39.6\% overall accuracy.
Further analysis shows that the three memory types fail for distinct reasons and that performance degrades as evidence spans longer temporal horizons, revealing that long-horizon memory remains far from solved.
We believe \benchmark{} establishes a strong foundation for evaluating and advancing long-context, memory-aware multimodal systems.
\end{abstract}

\section{Introduction}
\label{sec:intro}

Next-generation visual assistants, from smart glasses~\citep{grauman2022ego4d, Grauman_2024_CVPR,yu2026ego2web} to embodied agents~\citep{hu20253dllm, yang20253d, zhang2024vision} and always-on life-logging systems~\citep{xu2025streamingvlm}, must reason over continuous visual streams spanning an entire day or more.
This has driven growing interest in long-form video understanding~\citep{wang2025lvbench, Nagrani_2025_ICCV, yang2025cambrians, chandrasegaran2024hourvideo}, and more recently, in week-long video understanding ~\citep{yang2025egolife, chen2026towards, kim2026ma, yan2025teleego}.
At this temporal scale, relevant information is sparsely distributed across hours or days, posing unique challenges that go well beyond those of short-clip or hour-long video understanding.

\begin{figure}[t]
    \centering
    \includegraphics[width=\textwidth]{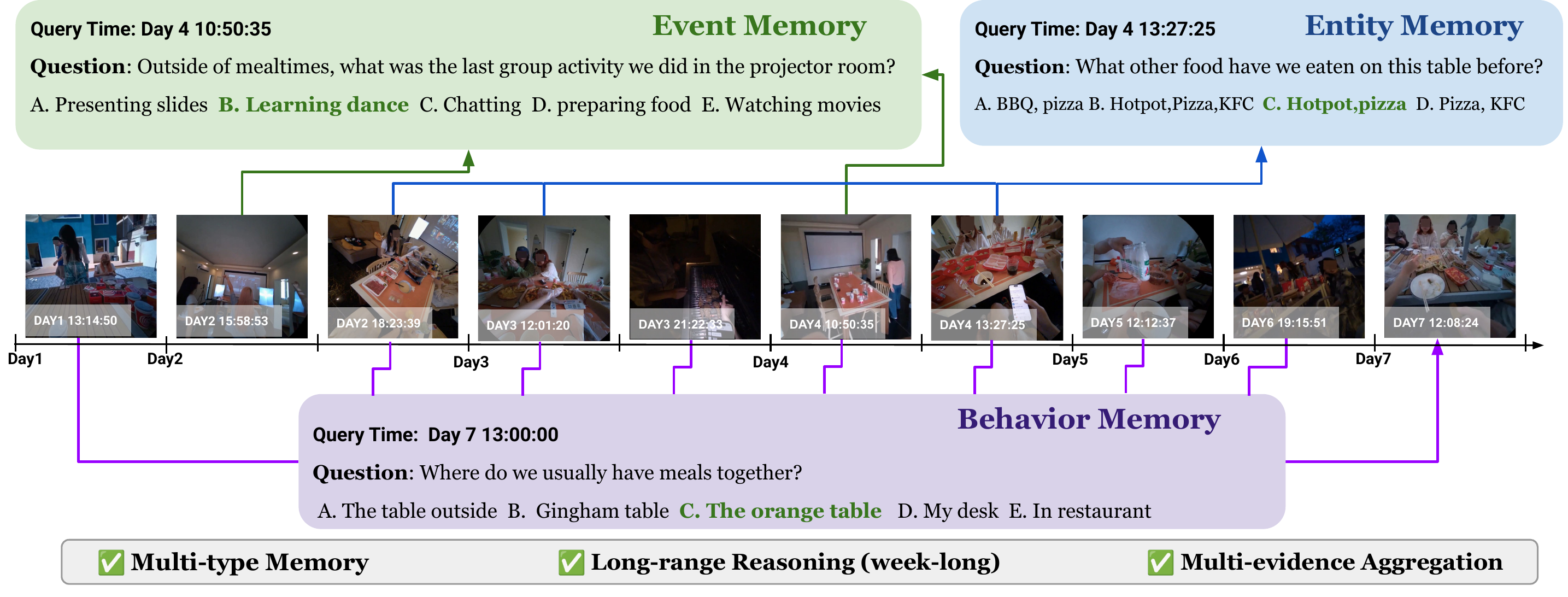}
    \caption{
        \textbf{Illustration of our \benchmark{} for week-long egocentric video memory.} Given a query at a specific time, answering requires retrieving and aggregating evidence from multiple temporally distant observations across days. We categorize memory into three types: entity memory (tracking persistent objects and states, tracking the same object for a long time), event memory (ordering and linking events,  e.g. linking previous similar event details), and behavior memory (inferring patterns, e.g. activity habit).
        }
    \label{fig:teaser_main}
    \vspace{-5mm}
\end{figure}

However, as video duration scales to days, densely sampling visual inputs (e.g., at 1 FPS) becomes impractical due to context length limitations, while the abundance of irrelevant content can further overwhelm models and hinder reasoning~\citep{liu2024lost}.
This makes \textbf{memory a fundamental challenge} for week-long video understanding: models must selectively accumulate information over time, recall previously observed states, track temporal order, and abstract recurring patterns from past experience.
Yet existing benchmarks~\citep{yang2025egolife, tian2025egor1chainoftoolthoughtultralongegocentric, yan2025teleego} primarily target perception and recognition; their questions are typically answerable from a single moment (e.g., ``What type of image did I paste into the slideshow?''~\citep{yan2025teleego}) or within a short temporal window of under ten minutes (e.g., ``What was in the pot just before it was set aside?''~\citep{yang2025egolife}), rather than requiring reasoning that accumulates multiple segments of evidence across hours to days. 
\begin{wrapfigure}{r}{0.45\textwidth}
    \centering
\includegraphics[width=0.45\textwidth]{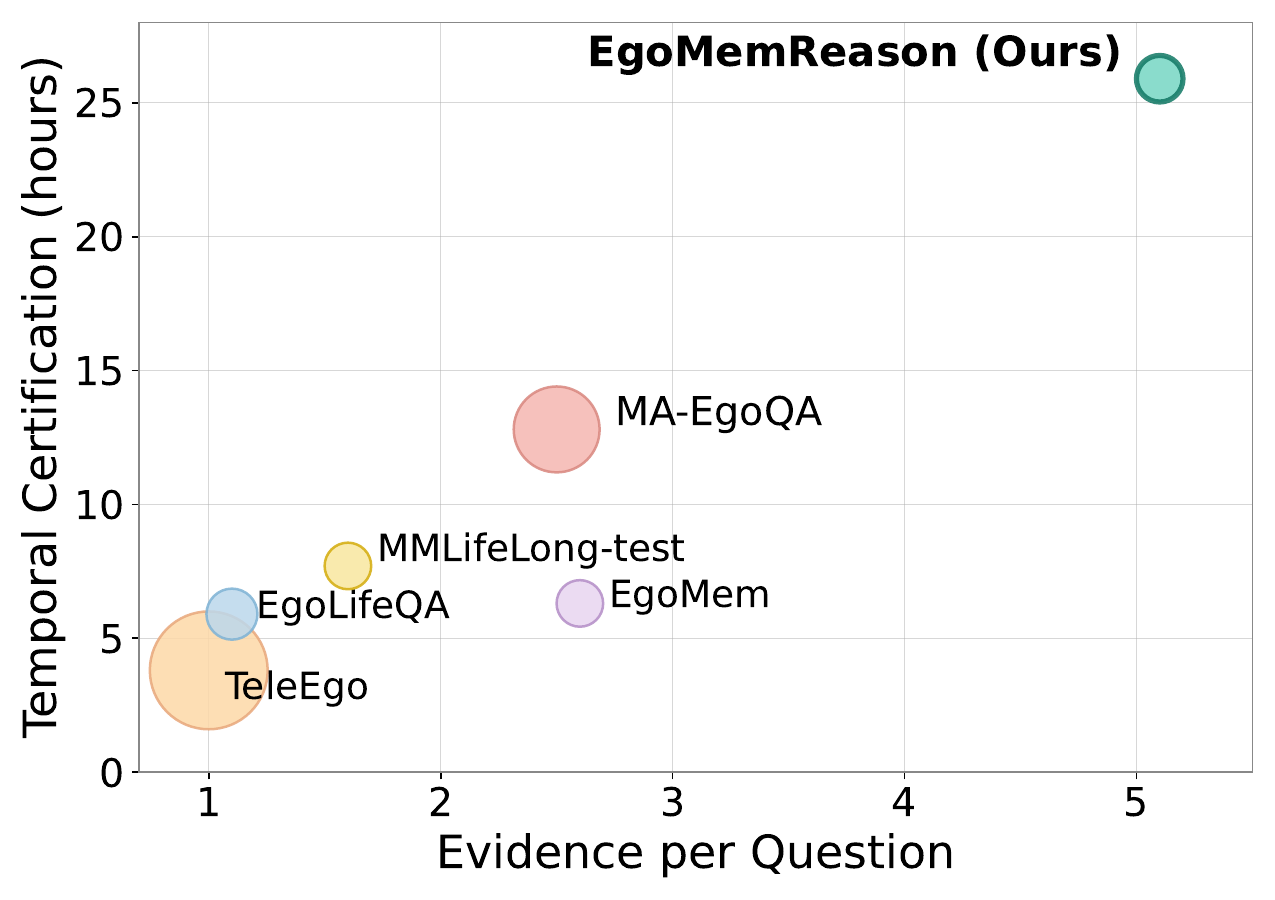}
    \caption{Comparison with existing week-long video benchmarks. The x-axis shows the average number of distinct video segments needed to answer a question (i.e., evidence), and the y-axis shows temporal certification in hours (i.e., the total video duration one must search to locate all ground-truth evidence). Bubble size is proportional to the number of questions.}
    \label{fig:comparision_benchmarks}
    \vspace{-5mm}
\end{wrapfigure}
As shown in~\Cref{fig:comparision_benchmarks}, most existing benchmarks span limited temporal certification of backtracking to the ground-truth evidence (defined in EgoSchema~\citep{mangalam2023egoschema}), and require only a small number of evidence segments per question.
Taken together, these limitations suggest that current benchmarks do not yet capture the memory demands of week-long video understanding, and a benchmark designed for this setting remains an open challenge.

To address this gap, we introduce \textbf{\benchmark{}}, a comprehensive benchmark designed to \textbf{systematically} evaluate week-long egocentric video understanding through the lens of memory-driven reasoning.
While existing long-video benchmarks largely reduce question answering to locating one or a few relevant moments and performing localized reasoning, as shown in ~\Cref{fig:teaser_main}, \benchmark{} requires models to aggregate, relate, and abstract over evidence distributed across days, operations that more closely resemble how humans actually reason over remembered experience. 
As shown in~\Cref{fig:comparision_benchmarks}, each question in \benchmark{} requires aggregating an average of $5.1$ distinct video segments distributed over average $25.9$ hours of memory backtracking, exceeding the strongest prior week-long benchmark by 2$\times$ in evidence count and 2$\times$ in temporal certification.
To capture the breadth of memory required at this temporal scale, as shown in \Cref{fig: task definition}, we decompose week-long memory into three complementary types, each targeting a distinct reasoning operation over accumulated experience.
\textit{Entity} memory, which requires \textbf{aggregating} how objects evolve across days (e.g., recalling all foods previously eaten at a particular table across multiple days); \textit{event} memory, which requires \textbf{relating} events separated by long intervals through ordering or linking (e.g., correctly sequencing activities spread across a week); and \textit{behavior} memory, which requires \textbf{abstracting} regularities from repeated experience (e.g., inferring where a person typically uses their phone based on accumulated observations). Together, these three types span the aggregation, relational, and inductive reasoning that genuine week-long memory entails, and that retrieval-centric evaluation cannot capture.

To construct \benchmark{}, we adopt a four-stage pipeline that combines \textbf{automated model-based generation with human verification}, transforming raw week-long egocentric video from EgoLife~\citep{yang2025egolife} into a rigorously verified question set.
We first convert each video into structured evidence through dense object-centric captioning and hierarchical event summarization with a strong MLLM.
We then design task-specific query generators that extract information from the week-long video and produce candidate multiple-choice questions, each constrained to a designated query timestamp so that only past observations are accessible.
Finally, all surviving candidates undergo human verification and revision 
using a multi-dimensional quality rubric assessing clarity, answer correctness, and option quality. Notably, annotators not only validate answers but also iteratively refine questions and distractors to eliminate ambiguity and strengthen visual grounding, resulting in 500 questions across six core challenges.

We evaluate \textbf{17} systems spanning three complementary paradigms: \textbf{8 general-purpose MLLMs} (Qwen-3-VL-8B, 30B-A3B and 32B~\citep{Qwen3-VL}, InternVL3.5-8B and 38B~\citep{wang2025internvl3_5}, GPT-5~\citep{openai2025gpt5}, Gemini-3-Flash~\citep{gemini3flash2025} and Gemini-3.1-Pro~\citep{googledeepmind2026gemini31pro}), \textbf{5 video-specific MLLMs} (LongVA~\citep{zhang2024longva}, InternVideo-2.5~\citep{wang2025internvideo}, VideoLLaMA3~\citep{damonlpsg2025videollama3}, Molmo2~\citep{molmo2openweightsdata}, StreamingVLM~\citep{xu2025streamingvlm}), and \textbf{4 agentic video frameworks} (SiLVR~\citep{zhang2026silvr}, Ego-R1~\citep{tian2025egor1chainoftoolthoughtultralongegocentric}, WorldMM~\citep{yeo2025worldmm}, AVP~\citep{wang2025active}).
Despite their model scale and pretraining, even the best model (Gemini-3-Flash) achieves only \textbf{39.6\%} overall accuracy.
Further analysis shows that the three memory types fail for fundamentally different reasons: entity memory is bottlenecked by fine-grained visual grounding, event memory by long-range temporal coherence, and behavior memory by abstraction over sparse repeated evidence, indicating that progress along three orthogonal axes is needed.
Ablation studies further show that neither much-denser frame sampling nor auxiliary text inputs (captions, transcripts) yield consistent improvement, suggesting that the challenge cannot be resolved simply through increased visual coverage or additional textual context.
Together, \benchmark{} and our analysis chart a path toward multimodal systems with structured, long-horizon memory that reasons beyond retrieval.

\section{Related Work} 

\noindent\textbf{Long Video Understanding (LVU).} 
Recent work has extended video understanding from short clips to long temporal horizons. Early benchmarks~\citep{tapaswi2016movieqa,lei2018tvqa,xiao2021next,wu2024star} focus on short videos with localized evidence, while newer datasets~\citep{mangalam2023egoschema,fu2025video,wu2024longvideobench,yang2025egolife,hu2025video,wang2025lvbench, tsuchiya2026ecbenchenumerationcountingbenchmark, hummel2024egocvr} evaluate longer videos with more complex reasoning~\citep{cheng2024egothink,chen2026towards,yan2025teleego, yu2026ego2web, chen2025grounded,chen2024cg}. A series of methods~\citep{yu2023self,yu2024frame,wang2024videoagent,zhang2023simple, tang2025adaptive,wu2019long,fan2024videoagent, goletto2024amego, he2024ma, song2024moviechat} study how to address LVU from different perspectives, including external memory~\citep{jin2025videomem, fan2024videoagent}, agentic pipeline~\cite{wang2025active,long2025seeing}, attention optimization~\citep{xu2025streamingvlm}, and so on. 
Recent benchmarks such as EgoLife~\citep{yang2025egolife}, MM-Lifelong~\citep{chen2026towards}, and MA-EgoQA~\citep{kim2026ma} move toward long-horizon, cross-event reasoning but differ in their treatment of memory. EgoLife, despite using ultra-long videos, relies on short-interval visual cues and lacks long-term temporal dependency. 
MM-Lifelong models long-term multimodal experience with an agentic memory mechanism, but primarily evaluates retrieval over long contexts. MA-EgoQA focuses on multi-agent interaction and shared context.  
Overall, most LVU benchmarks remain retrieval-centric, reducing tasks to locating a few relevant segments and performing localized reasoning, without systematically studying memory mechanisms. 
In contrast, we explicitly study \emph{structured memory}, where evidence is distributed over long time spans and must be incrementally accumulated before retrieval becomes meaningful.

\noindent\textbf{Memory Benchmarks in Text and Multimodal Domains.} 
Recent work evaluates long-term memory across text and multimodal settings. 
In text, synthetic benchmarks~\citep{hsieh2024ruler, kuratov2024babilong} offer controlled evaluation but rely on artificial signals, while task-driven~\citep{hu2025evaluating,wang2025memalpha}, conversational~\citep{maharana2024lococmo,wu2024longmemeval}, and narrative benchmarks~\citep{wan2025storybench,kim2026can} require tracking information over extended contexts. In multimodal domains, memory is increasingly critical as information is distributed across time and modalities. Prior work in long-form and egocentric video~\citep{yang2025egolife,chen2026towards, zhou-etal-2025-x, perrett2025hd, Baermann_2022_CVPR, 10.1007/978-3-032-10192-1_42}, multimodal dialogue~\citep{bei2026mem}, and embodied systems~\citep{datta2022episodic,savva2019habitat, zhang2024vision} requires integrating visual, linguistic, and interaction histories over long horizons. 
However, these benchmarks are primarily \emph{task-driven} and do not explicitly isolate the structure of memory required for success, leaving unclear what information is stored, how it is updated, and which memory types models rely on.
\benchmark{} addresses this gap by explicitly decomposing long-horizon video memory into three functionally distinct types: entity, event, and behavior, which capture object-state trajectories, temporally grounded episodes, and patterns abstracted from repeated observations, respectively.

\section{Benchmark Construction}
\label{sec:benchmark}

Our benchmark is organized around three complementary memory types and six core challenges in total.
Every question is designed to demand reasoning over accumulated temporal evidence across the long video, rather than single-clip retrieval or surface-level pattern matching.
Formally, given an observed video frame sequence $V_{\le t_q}=\{v_1,\dots,v_{t_q}\}$ up to query time $t_q$ and a query $q$, a model $f_\theta^{(m)}$ produces an answer
\begin{equation}
\hat{y}^{(m)} = f_\theta^{(m)}\!\left(q,\, \mathcal{M}^{(m)}_{\le t_q}(V_{\le t_q})\right),
\end{equation}
where $m\in\{\mathrm{entity},\mathrm{event},\mathrm{behavior}\}$ denotes the memory type, $\mathcal{M}^{(m)}_{\le t_q}$ is the corresponding structured memory built from past observations $V_{\le t_q}$, and $\hat{y}^{(m)}$ is the predicted answer.
In the following, we present the memory design principle (\S\ref{sec:taxonomy}), task definitions for each of the six challenges (\S\ref{sec: task definition}), and the four-stage data generation pipeline together with benchmark statistics (\S\ref{sec:main pipeline}). More detailed formulations for each task are provided in Appendix~\ref{supp:pipeline}.

\subsection{\benchmark{} Design Principle}
\label{sec:taxonomy}
Cognitive science has long established that human memory is not a monolithic retrieval store but comprises qualitatively distinct systems~\citep{Kahneman1992TheRO, tulving1972episodic}, each serving a different functional role: \textbf{entity memory} allows individuals to track and re-identify objects and people as they change over time, \textbf{event memory} supports the recollection of specific experienced events situated in time and place, and \textbf{behavior memory} enables the abstraction of general knowledge detached from any particular episode. 
These categories are complementary: each operates at a different granularity and timescale, and together they enable flexible, efficient, and robust cognitive performance. 
This principle has also been adopted in LLM-based agents, which increasingly incorporate structured memory modules inspired by these cognitive categories~\citep{park2023generative_agents,zhang2024survey_agent}. 
Recent benchmarks~\citep{wu2024longmemeval, maharana2024lococmo, wan2025storybench, hu2025evaluating} probe distinct memory competencies such as retention, retrieval, and update, but examine these capabilities in isolation without considering how different memory types interact. 
This limitation is even more pronounced in multimodal settings, where existing evaluations rarely distinguish memory types or require their integration over long temporal horizons. 
Evaluating these capabilities in a structured and disentangled manner is therefore critical for diagnosing model limitations in long-horizon video understanding.

\begin{figure}[t]
    \centering
    \includegraphics[width=\linewidth]{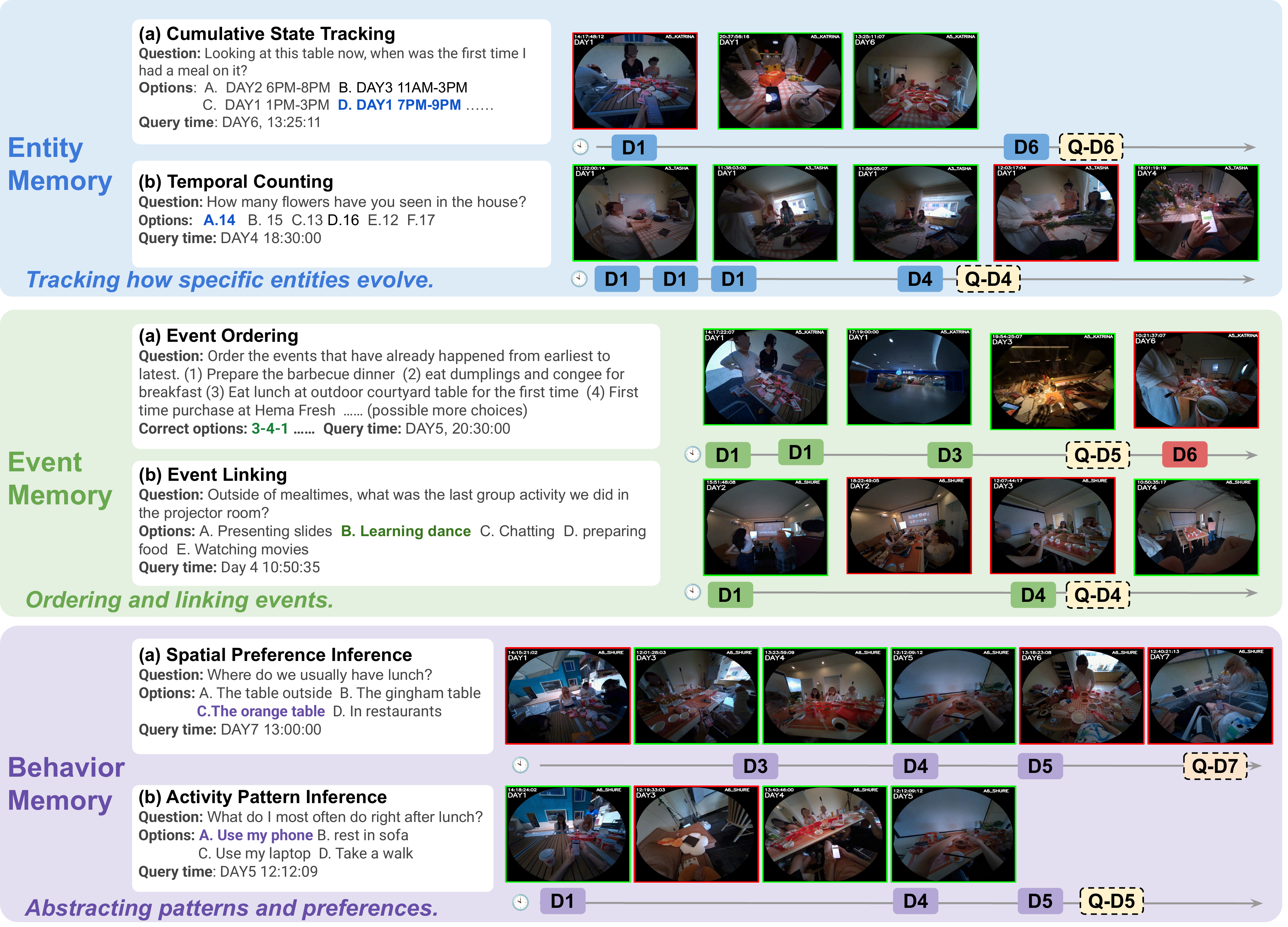}
\caption{Overview of the six core challenges across three memory types in \benchmark{}. Within each example, the week-long timeline shows evidence frames sampled at different timestamps (e.g., \textbf{D1 and D4 denote evidence days, and Q-D6 indicates the query timestamp on Day 6, highlighted by a dashed box}). The timeline provides a unified temporal layout over a week and does not necessarily correspond one-to-one with the frames shown above. Green frames indicate relevant evidence; red frames indicate distractors that should not contribute to the answer — either because they are unrelated to the query, or because they occur \emph{after} the query timestamp (e.g., D6 in Event Ordering lies beyond the Day-5 query and is therefore inadmissible evidence).
    }
    \vspace{-5mm}
    \label{fig: task definition}
\end{figure}

\subsection{\benchmark{} Task Definition}
\label{sec: task definition}
We formulate long-horizon video understanding as a \emph{structured memory reasoning} problem. Building on the three memory types defined in~\S\ref{sec:taxonomy}, we operationalize each into two tasks targeting distinct reasoning demands, yielding six tasks in total (\Cref{fig: task definition}). Each task is designed so that answering requires aggregating evidence from multiple temporally distributed observations rather than retrieving a single moment.

\noindent\textbf{Entity Memory.} At week-long scales, entities appear, disappear, and resurface in altered states---under different lighting, from new viewpoints, or in entirely different locations. Re-identifying them and tracking how they evolve over such intervals demands more than single-frame recognition. We evaluate this through two core capabilities:
\begin{enumerate}[nosep,leftmargin=1.5em]
  \item \emph{Cumulative State Tracking.}
  Given an entity observed at multiple points in the video, the task is to identify how its location or condition has changed across observations separated by hours or days. As shown in the cumulative state tracking example of~\Cref{fig: task definition}, the model may need to track that ``a bowl'' initially appears on the ``kitchen counter'', is later moved to the ``sink'', and finally placed on the ``dining table'', with each observation separated by hours.

  \item \emph{Temporal Counting.} This task requires reasoning over sets of entities by counting how many distinct instances of a category have appeared across the video. 
Unlike static counting, the count is defined with respect to a query timestamp, such that only instances observed up to that time are considered.
This requires constructing a global inventory of entity instances over time, identifying repeated occurrences under varying visual conditions, and distinguishing instances that may share similar appearance or context. 
\end{enumerate}

\noindent\textbf{Event Memory} evaluates the ability to retrieve, temporally organize, and relate discrete events from the video history.
Week-long videos contain a rich stream of activities that unfold over hours or days, where later events often depend on, revisit, or modify earlier ones.
We evaluate event memory through two core capabilities:

\begin{enumerate}[nosep,leftmargin=1.5em]
  \item \emph{Event ordering.}  Given a set of events drawn from different days, the task requires arranging them in correct temporal order.
  This demands maintaining a structured timeline of past activities across large temporal gaps.
 
  \item \emph{Event Linking.} Given a set of contextual constraints (e.g., location, activity type, or time of day), the task requires identifying the relevant event matching those conditions from the previous video inputs. 
This requires reasoning over hours or even days, filtering candidates under multiple constraints, and selecting the correct event among visually and semantically similar alternatives.
\end{enumerate}

\noindent\textbf{Behavior Memory} tests whether a system can abstract higher-level knowledge from repeated observations over time, going beyond individual-event recall to form generalized priors.
While entity memory focuses on specific objects and event memory on specific events, behavior memory requires distilling patterns that no single observation can reveal.
We evaluate behavior memory through two core capabilities:
\begin{enumerate}[nosep,leftmargin=1.5em]
  \item \emph{Spatial Preference Inference.}
  This task requires inferring recurring patterns or habitual associations from repeated observations over time, such as spatial preferences (e.g., where a person typically performs a given activity) or common activities at a location.
  \item \emph{Activity Pattern Inference.}
This task requires predicting likely next states based on learned behavior patterns, such as the next location given the current one or the next activity given the current context (e.g., ``Where is the person most likely to go after lunch?'').
  These questions test whether the system has internalized patterns from the daily routines.
\end{enumerate}

\begin{figure}
    \centering
    \includegraphics[width=\linewidth]{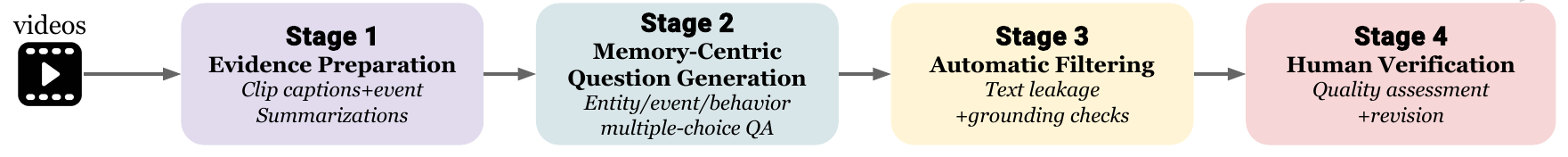}
\caption{Overview of the benchmark construction pipeline. \textbf{Stage 1} extracts clip captions and event summaries from raw egocentric videos; \textbf{Stage 2} generates entity/event/behavior multiple-choice questions from this evidence; \textbf{Stage 3} filters text-leakage cases and verifies temporal grounding; \textbf{Stage 4} performs human quality assessment and revision.}    \label{fig:benchmark pipeline}
\end{figure}

Under this unified formulation, all tasks require retrieving relevant observations from long video histories, but differ in the structure of memory to be constructed: entity memory tracks persistent state trajectories, event memory organizes temporally grounded episodes, and behavior memory abstracts recurring patterns across time. This distinction enables controlled evaluation of how models store, update, and reason over long-horizon multimodal information. Together, the three memory types and their six core capabilities provide a comprehensive evaluation of the memory capacities required for long-term memory in week-long egocentric video understanding.

\subsection{Benchmark Construction Pipeline and Statistics}
\label{sec:main pipeline}
Our benchmark is built on videos from the EgoLife dataset~\citep{yang2025egolife}, which provides week-long, continuous egocentric recordings across six 
participants in naturalistic daily routines. The multi-day, always-on nature of these videos, capturing recurring activities, evolving object states, and extended social interactions, makes them uniquely suited for evaluating long-horizon memory.
\begin{wrapfigure}{r}{0.35\textwidth}
    \centering
\includegraphics[width=0.35\textwidth]{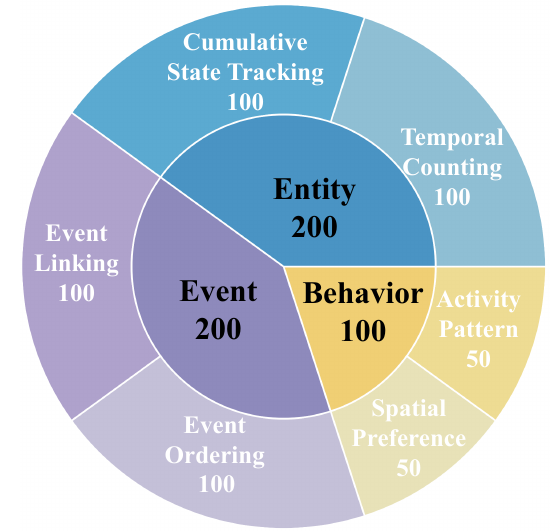}
    \caption{Dataset composition by memory type.}
    \label{fig:data_distribution}
\end{wrapfigure}
\benchmark{} is formulated as a multiple-choice question-answering benchmark, with each question paired with one correct answer and several semantically competitive distractors.
As shown in \Cref{fig:benchmark pipeline}, we construct questions through a four-stage pipeline designed to ensure that every question is temporally grounded, visually verified, and genuinely challenging.
\Cref{fig:data_distribution} summarizes the composition of \benchmark{}, which contains 500 questions across three memory types and six core challenges. Further details on the construction pipeline are provided in Appendix~\ref{app: Benchmark Details and Statistics}.
 
\noindent\textbf{Stage 1: Evidence Preparation.}
\label{sec:stage1}
We convert raw multi-day egocentric video into structured evidence by generating both fine-grained clip-level captions and higher-level event summaries. 
Videos are segmented into non-overlapping 30-second clips and processed with GPT-5~\citep{openai2025gpt5} to produce object-centric descriptions, tracking state changes (e.g., open/closed, on/off, and full/empty), spatial locations, and human interactions (e.g., who holds, uses, or hands off the object) as well as event-level activities and their temporal context.
These dense captions are further organized into a hierarchical event structure spanning three levels of temporal granularity (30-second, 10-minute, 2-hour, and day-level), annotated with activity labels, location tags, and object references, serving as the primary signal for the query generation.

\noindent\textbf{Stage 2: Query Generation.}
\label{sec:stage2}
From the structured evidence, we generate candidate multiple-choice questions for each of the three memory types, each associated with a query time where only prior observations are accessible. We use GPT-5.2~\citep{openai2025gpt5}, a different version within the same model family as the captioning stage to avoid self-reinforcing biases.
For each memory type, we design a task-specific pipeline comprising three steps: (1) \emph{statement extraction}, which identifies and aggregates relevant factual statements from the structured evidence to serve as the basis for question formulation; (2) \emph{query generation}, which formulates questions targeting the corresponding memory capability; and (3) \emph{distractor generation}, which produces semantically competitive incorrect options drawn from similar contexts to ensure non-trivial difficulty.

\noindent\textbf{Stage 3: Automatic Filtering.}
\label{sec:stage3}
We apply model-based filtering to remove trivial, ambiguous, or ungrounded questions.
A blind test presents each question without visual input to three LLMs (Gemini-3.1-Pro~\citep{googledeepmind2026gemini31pro}, GPT-5~\citep{openai2025gpt5}, Qwen-3-VL-32B~\citep{Qwen3-VL}); questions answered correctly by a majority are discarded.
We then verify that correct answers are supported by valid visual evidence before the query time, enforce a minimum temporal gap of 2 hours across supporting evidence, well beyond existing benchmarks~\citep{fu2025video, Zhou2024MLVUBM, wang2025lvbench}.
Details are provided in Appendix~\ref{supp:pipeline}.

\noindent\textbf{Stage 4: Human Verification.}
\label{sec:stage4}
All remaining candidates undergo human verification through a dedicated annotation interface that presents annotators with the query-time context video and pre-selected evidence clips alongside each question. Six annotators at the college or graduate level review each surviving question, spending approximately 20 minutes per sample to assess (1) question clarity, (2) answer correctness, and (3) option quality, and can revise or reject problematic samples. This process forms a quality-control loop in which revised questions are re-validated before inclusion, ensuring that the final benchmark is both visually grounded and human-validated. Overall, only 15\% of initial candidates survive the combined model-based filtering and human verification stages, reflecting the stringent quality standards applied throughout the pipeline.
We provide detailed human verification and UI in ~\Cref{fig:annotation-ui} in the Appendix. 

\section{Experimental Results}
\label{exp:benchmark}

\subsection{Experimental Setup}
\noindent\textbf{Evaluated Models.}
We evaluate 17 systems spanning three complementary paradigms to assess how different architectural and reasoning strategies handle memory-intensive, long-horizon tasks. This includes general-purpose MLLMs (Qwen-3-VL~\citep{Qwen3-VL}, InternVL3.5~\citep{wang2025internvl3_5}, GPT-5.2~\citep{openai2025gpt5}, Gemini-3-Flash~\citep{gemini3flash2025}, Gemini-3.1-Pro~\citep{googledeepmind2026gemini31pro}) covering a wide range of model scales and capabilities; video-specific MLLMs (LongVA~\citep{zhang2024longva},  StreamingVLM~\citep{xu2025streamingvlm}), InternVideo-2.5~\citep{wang2025internvideo}, VideoLLaMA3~\citep{damonlpsg2025videollama3}, Molmo2~\citep{molmo2openweightsdata}) incorporating extended temporal modeling or video-centric pretraining; and agentic video frameworks (SiLVR~\citep{zhang2026silvr}, Ego-R1~\citep{tian2025egor1chainoftoolthoughtultralongegocentric}, WorldMM~\citep{yeo2025worldmm}) and AVP~\citep{wang2025active} that decompose reasoning into structured sub-tasks with retrieval or external memory. 

\noindent\textbf{Evaluation Metric.}
We report standard multiple-choice question accuracy (\%) at multiple granularities: per capability (cumulative state tracking, temporal counting, event ordering, event linking, spatial preference inference, activity pattern inference), per memory type (Entity, Event, Behavior) and overall performance.

\noindent\textbf{Implementation Details.}
For all general MLLMs and VideoLLMs, as dense 1FPS sampling from week-long video is impractical for all baselines, we uniformly sample 1024 frames
from the video start to the query timestamp $t_q$, which are further down-scaled if necessary to fit each model's context length. 
For agentic video frameworks, we adopt each method's best-performing configuration reported on EgoLife~\citep{yang2025egolife}; when the original settings are not accessible, we default to GPT-5~\citep{openai2025gpt5} as the main agent. 
More implementation details are included in Appendix~\ref{more implementationd details}.

\subsection{Main Results}

\begin{table*}[t]
\centering
\scriptsize
\label{tab:main_results}
\resizebox{0.95\textwidth}{!}{%
\begin{tabular}{l|cc|cc|cc|c}
\toprule
\multirow{2}{*}{\textbf{Method}} 
& \multicolumn{2}{c|}{\textbf{Entity Memory}} 
& \multicolumn{2}{c|}{\textbf{Event Memory}} 
& \multicolumn{2}{c|}{\textbf{Behavior Memory}} 
& \multirow{2}{*}{\textbf{Overall}} \\
\cmidrule(lr){2-3} \cmidrule(lr){4-5} \cmidrule(lr){6-7}
& Tracking & Counting & Ordering & Linking & Spatial & Activity & \\
\midrule
\rowcolor{gray!10}
Random             & $19.6$ & $16.7$ & 11.1 & $17.3$ & $19.3$ & $19.2$ & 16.8  \\

\midrule
\multicolumn{8}{l}{\textit{\textbf{General MLLMs}}} \\
\midrule
InternVL3.5-8B~\citep{wang2025internvl3_5}     & 23.0 & 29.0 & 23.0 & 27.0 & 34.0 & 42.0 & 28.0 \\

Qwen-3-VL-8B~\citep{Qwen3-VL}       & $35.0$ & $28.0$ & 23.0 & $21.0$ & $40.0$ & $42.0$ & 29.6 \\
InternVL3.5-38B~\citep{wang2025internvl3_5}    & 33.0 & 40.0 & 27.0 & 24.0 & 46.0 & 32.0 & 32.6\\
Qwen-3-VL-30B-A3B~\citep{Qwen3-VL}      &36.0 & \underline{48.0} & 25.0 & 26.0 & 40.0 & 30.0 & 34.0 \\
Qwen-3-VL-32B~\citep{Qwen3-VL}      & 35.0 & {46.0} & 28.0 &  27.0 & \textbf{50.0} & \underline{46.0} & 36.8 \\
GPT-5    ~\citep{openai2025gpt5}          & 29.0 & 42.0 & 20.0   & 18.0 &32.0  & 28.0      &  27.8  \\ 
Gemini-3-Flash~\citep{gemini3flash2025}     & \textbf{46.0} & 28.0 & \underline{36.0} & \textbf{44.0} & 44.0 & 44.0 &  \textbf{39.6} \\
Gemini-3.1-Pro~\citep{googledeepmind2026gemini31pro}     & \underline{40.0} & 26.0 &  \textbf{44.0} & \underline{33.0}   & 40.0   & \textbf{48.0}   & \underline{37.4} \\
\midrule
\multicolumn{8}{l}{\textit{\textbf{Video-specific MLLMs}}} \\
\midrule
LongVA-7B~\citep{zhang2024longva}          & 22.0 & 18.0 & 20.0 & 22.0 & 20.0 & 22.0 & 20.6 \\
StreamingVLM~\citep{xu2025streamingvlm}   & 25.0 & 29.0 & 21.0 & 20.0 & 20.0 & 32.0 & 24.2 \\
InternVideo-2.5-8B~\citep{wang2025internvideo} & 29.0 & 27.0 & 25.0 & 15.0 & 32.0 & 32.0 & 25.6 \\
VideoLLaMA3-8B~\citep{damonlpsg2025videollama3}     & 23.0 & 31.0 & 27.0 & 32.0 & 38.0 & 36.0 & 30.0 \\
Molmo2-8B~\citep{molmo2openweightsdata}          & 36.0 & \textbf{50.0} & 27.0 & $25.0$ & 34.0 & $22.0$ & 33.2\\
\midrule
\multicolumn{8}{l}{\textit{\textbf{Agentic Video Frameworks}}} \\
\midrule
SiLVR~\citep{zhang2026silvr}              & 31.0 & 14.0  & 27.0 & 17.0 & 18.0 & 28.0 & 22.4 \\
Ego-R1~\citep{tian2025egor1chainoftoolthoughtultralongegocentric}        & 30.0 & 18.0 & 23.0 & 18.0 & \underline{48.0} & 32.0 & 25.8 \\
WorldMM~\citep{yeo2025worldmm}            & 32.0 & 44.0 & 21.0  & 21.0 & 34.0 & 36.0 & 30.6 \\
AVP ~\citep{wang2025active}                & 34.0 & 42.0 &  31.0  & 27.0 & 38.0   & 34.0   &  34.0 \\
\bottomrule
\end{tabular}
}
\caption{Main benchmark results on \benchmark{}. We report accuracy (\%) across three memory types and six capability dimensions: Tracking (Cumulative State Tracking), Counting (Temporal Counting), Ordering (Event Ordering), Linking (Event Linking), Spatial (Spatial Preference Inference), and Activity (Activity Pattern Inference). The best result in each column is \textbf{bolded} and the second best is \underline{underlined}.}
\vspace{-5pt}
\end{table*}

In \Cref{tab:main_results}, we summarize the performance of all evaluated systems on \benchmark{} and highlight several key findings.
Across all evaluated models, performance remains low on \benchmark{}, with no model achieving strong accuracy across all memory types. Results vary significantly across models and tasks, indicating that long-horizon memory reasoning remains an open challenge.

\textbf{No single approach consistently leads.}
The best overall accuracy is Gemini-3-Flash~\citep{gemini3flash2025} at 39.6\%, with no model reaching 51\% on any single capability.
Scaling model size within families yields moderate gains: Qwen-3-VL improves by 7.2\% from 8B to 32B, and InternVL3.5 gains 4.6\% from 8B to 38B, suggesting that increasing parameters alone does not resolve the core challenges of long-horizon memory reasoning.
Across paradigms, no category consistently outperforms the others: general MLLMs lead on Event and Behavior memory, with Gemini-3.1-Pro topping Event Ordering, Gemini-3-Flash topping Event Linking, and Qwen-3-VL-32B topping Spatial Preference Inference; video-specific MLLMs are competitive on Entity memory, where Molmo2-8B achieves the highest Temporal Counting score across all 17 systems, reflecting the benefit of pixel-level grounding pretraining; and agentic frameworks generally underperform general MLLMs, likely for two reasons: (i) most rely on caption-based intermediate representations that lose fine-grained visual cues critical for Entity Memory, and (ii) they are primarily designed for and evaluated on hour-scale videos, leaving them under-equipped for the week-long evidence aggregation required by \benchmark{}.
Within these limits, agentic frameworks still show targeted strengths, with Ego-R1 reaching 48.0\% on Spatial Preference Inference where retrieving a single dominant pattern suffices.

\textbf{Different memory types expose different bottlenecks.}
The results show that the three memory types fail for distinct reasons, each pointing to a different missing capability rather than a shared limitation.
\emph{Entity Memory} is bottlenecked by fine-grained visual grounding combined with long-context modeling: models that rely more heavily on text-centric reasoning or training (e.g., LongVA, SiLVR) fall below 25\% on Temporal Counting, while Molmo2-8B, which combines pixel-level grounding pretraining (e.g., pointing and tracking) with long-context post-training, leads all 8B models on both Cumulative State Tracking and Temporal Counting, indicating that Entity Memory benefits from both perceptual precision and the ability to retain visual evidence over extended temporal contexts.
\emph{Event Memory} is bottlenecked by long-range temporal coherence: even the strongest models get 44\% accuracy on both Event Ordering and Event Linking examples.
This contrasts sharply with prior egocentric benchmarks~\citep{yang2025egolife, yan2025teleego}, where models achieve much higher accuracy on single-event retrieval, suggesting that current models can locate individual events but struggle when evidence must be related across extended temporal spans, a pattern further confirmed by the sharp drop in Event accuracy as temporal certification length grows~(\Cref{tab:cert_length}).
\emph{Behavior Memory} is bottlenecked by long-horizon reasoning over sparse repeated evidence: even the best models stay at 50.0\% on Spatial Preference Inference and 48.0\% on Activity Pattern Inference.
This contrasts with prior long-video benchmarks~\citep{yan2025teleego, mangalam2023egoschema}, where models achieve strong performance on global video summarization, suggesting that current models can summarize what they have seen but struggle to abstract recurring patterns across many sparsely distributed observations.
Together, these gaps confirm that progress on long-horizon video understanding requires advances on three orthogonal axes: perceptual precision combined with long-context retention for entities, structured temporal modeling for events, and aggregation-based reasoning for behaviors, none of which are addressed by simply scaling model size or input length.

\subsection{Ablation Studies and Analysis}
We further conduct a series of analyses to better understand the underlying bottlenecks revealed in the main results, focusing on the roles of temporal length, visual input scaling, and auxiliary information.
Unless otherwise specified, all experiments are conducted using Qwen-3-VL-8B~\citep{Qwen3-VL}.
In addition, we provide further ablation studies on prompt strategies, additional analysis in Appendix~\ref{supp:more_experimental_results}. 

\begin{figure}[t]
    \centering

    \begin{minipage}[t]{0.5\linewidth}
        \vspace{12pt}
        \centering
        \scriptsize
        \resizebox{\linewidth}{!}{
        \begin{tabular}{l|ccccc}
        \toprule
        \textbf{Cert. Length (h)} & $<$8 & 8--16 & 16--32 & 32+ & Total \\
        \midrule
        Entity       & 28.5 & 33.9 & 32.1 & 30.3 & 31.5 \\
        Event        & --   & 31.1 & 23.0 & 13.5 & 22.0 \\
        Behavior   & --   & --   & 43.7 & 37.0 & 41.0 \\
        \midrule
        Overall      & 40.3 & 33.7 & 32.5 & 23.2 & 29.6 \\
        \bottomrule
        \end{tabular}
        }
        \captionof{table}{Effect of temporal certification on accuracy (\%) across memory types. ``--'' indicates bins with no examples for the corresponding memory type, as those questions occur only at longer temporal spans.}
        \label{tab:cert_length}
    \end{minipage}
    \hfill
    \begin{minipage}[t]{0.42\linewidth}
        \vspace{0pt}
        \centering
        \includegraphics[
            width=\linewidth,
            trim=0 0 0 0,
            clip
        ]{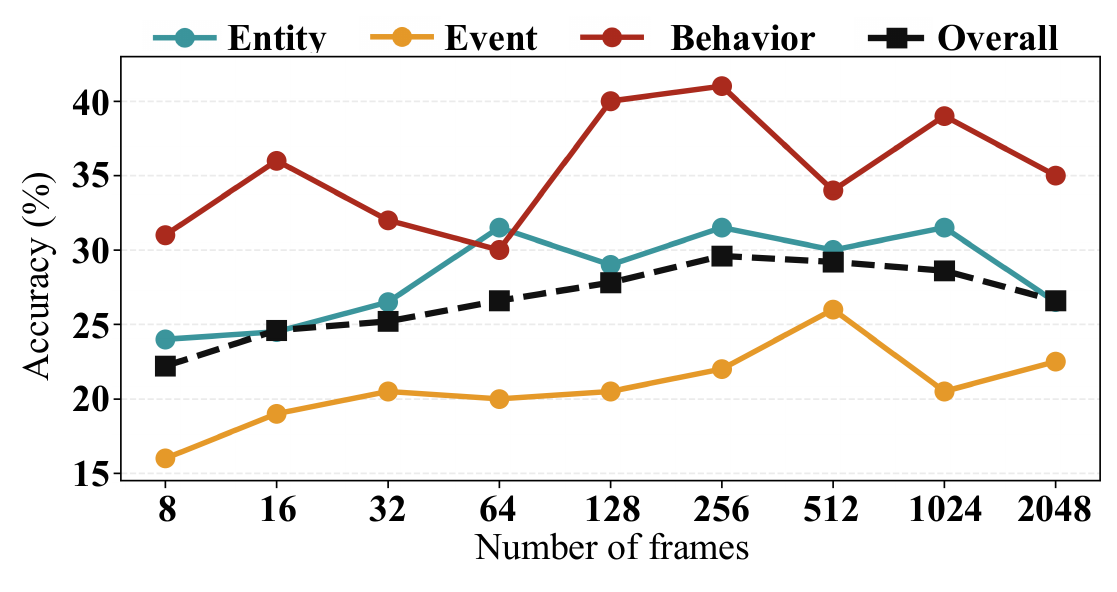}
        \captionof{figure}{Effect of input frame count on accuracy across
        memory types.}
        \label{fig:frame_ablation}
    \end{minipage}

    \vspace{-10pt}
\end{figure}

\noindent\textbf{Effect of temporal certification.}
In Table~\ref{tab:cert_length}, we analyze performance as a function of temporal certification length, defined as the total video duration one must search to locate all ground-truth evidence. 
Overall accuracy decreases as the temporal span increases, confirming that longer evidence spans pose a substantial challenge regardless of memory type.
The three memory types exhibit distinct degradation patterns. 
Event memory shows the sharpest and most monotonic decline, falling by more than half across the available ranges, indicating that event memory is the most temporally sensitive of the three.
Event memory shows the sharpest decline, dropping substantially as the evidence span grows from short to medium ranges and continuing to fall at the longest spans.
Behavior memory is only defined for longer spans and declines moderately as the temporal window extends.
Together, these patterns show that the impact of temporal span varies substantially by memory type, indicating that benchmarks evaluating long-horizon understanding must consider per-type temporal dynamics rather than reporting a single aggregate trend.

\noindent\textbf{Frame input scaling.}
We analyze how performance changes as the number of sampled input frames increases, as shown in ~\Cref{fig:frame_ablation}.
Entity memory improves steadily with more frames before saturating around 256, indicating that visual coverage helps up to a point but cannot compensate for models' limited capacity to extract fine-grained object states from individual frames.
Event memory is the least responsive to frame scaling, which suggests that the bottleneck is not visual coverage but the inability to maintain long-range temporal coherence when ordering and linking events across days.
Behavior memory benefits most from denser sampling but remains highly unstable across budgets, reflecting its dependence on capturing recurring patterns, a signal that is easily disrupted when additional frames introduce conflicting or off-routine observations.
Overall, no single frame budget is optimal across different memory types, indicating that scaling frame count alone cannot address long-horizon memory.

\paragraph{Quantitative error analysis.} To better understand the limitations of current MLLMs, following existing work~\citep{cheng2025video}, we randomly sampled 
100 benchmark examples and manually inspected the failure cases. Our analysis on Gemini-3-Flash reveals four common failure patterns: (1) extracting incorrect visual details from the video (28\%), (2) missing important visual information, especially in long-range reasoning scenarios (32\%), (3) perceiving the correct visual details but making logical mistakes during reasoning (32\%), and (4) producing incorrect predictions despite correct reasoning processes (8\%).

\section{Conclusion}
We introduced \benchmark{}, a benchmark for evaluating long-horizon memory reasoning in week-long egocentric videos, decomposing memory into three complementary types (entity, event, and behavior) across six core challenges that require multi-hop reasoning over temporally distributed evidence.
Evaluating 17 systems reveals that long-horizon memory remains a substantial open challenge: the three memory types fail for fundamentally different reasons. Entity memory is bottlenecked by fine-grained visual grounding over long time period, event memory by long-range temporal coherence, and behavior memory by abstraction over sparse repeated evidence.
We further show through ablations on temporal length, frame input, and auxiliary information that none of these bottlenecks can be addressed simply by scaling input or context.
We believe \benchmark{} serves as a rigorous diagnostic framework for guiding future research toward models capable of genuine long-horizon memory reasoning.

\section*{Acknowledgment}
We would like to thank David Wan, Nithin Sivakumaran, and Fengli Wu for their help in the human annotation process. 
This work was supported by ONR Grant N00014-23-1- 2356, ARO Award W911NF2110220, DARPA ECOLE Program No. HR00112390060, NSF-AI Engage Institute DRL-2112635, Laboratory for Analytic Sciences via NC State University, National Institutes of Health Award 1R01HD111074-01, and Sony Focused Research award. 
The views contained in this article are those of the authors and not of the funding agency.

\bibliography{colm2026_conference}
\bibliographystyle{colm2026_conference}

\appendix
\newpage
\section*{Appendix}

In this appendix, we first describe the full data construction pipeline (\S\ref{supp:pipeline}).
We then present additional implementation details (\S\ref{more implementationd details}) and additional analysis (\S\ref{supp:more_experimental_results}).
Finally, we discuss limitations and directions for future work (\S\ref{limitations}).

\section{Data Construction Details}
\label{supp:pipeline}

Our benchmark is built on videos from the EgoLife dataset~\citep{yang2025egolife}, which provides ultra-long, continuous egocentric recordings spanning multiple days across six participants engaged in naturalistic daily routines.
This multi-day, always-on nature of the video makes it uniquely suited for evaluating long-horizon memory: the recordings capture rich temporal dynamics, including recurring activities, evolving object states, and extended social interactions that unfold across days rather than minutes.

As shown in previous Figure~\ref{fig:benchmark pipeline}, we construct questions through a four-stage pipeline designed to ensure that every retained item is temporally grounded, visually verified, and genuinely challenging.

\subsection{Stage 1: Evidence Preparation}
\label{app:stage1}
The first stage converts raw multi-day egocentric video into structured textual evidence that supports downstream question generation. We segment each participant's recording into 30-second clips and caption them with a VLM following a structured, object-centric rubric. These clip-level captions are then aggregated into event-level summaries annotated with activity labels, location tags, and object references. The resulting dual-granularity representation, consisting of fine-grained clip timelines paired with coarser event scaffolds, serves as the primary input for question generation.

\paragraph{Clip-level Captioning.}
We segment each participant's recording into non-overlapping 30-second clips and sample frames within each clip at 1~FPS. A VLM (GPT-5) receives the sampled frames and produces object-centric state-tracking annotations following a structured rubric. 
The rubric directs the model to attend to five dimensions in order: presence, including appearance, disappearance, addition, and removal; attribute or status changes such as open/closed, on/off, full/empty, or clean/dirty; spatial location and movement; interaction with people, covering who holds, uses, or hands off each object; and count or accumulation over time.
For each clip, the model produces a dense caption that records, where applicable, each object's current status, location, interaction context, what changed relative to earlier in the clip, and the specific evidence frames. This object-centric design grounds the captions in observable states rather than narrative summaries, providing the fine-grained temporal evidence needed for downstream entity-tracking and episodic-memory questions.

\paragraph{Hierarchical Event Summarization.}
Beyond clip-level captions, we summarize each participant's recording at multiple temporal scales along a fixed pyramid: 30-second clips, 10-minute windows, 2-hour windows, and full-day summaries.
Each level is produced by prompting GPT-5 to summarize the captions from the level below into a progressively coarser description that retains activity, location, and object references.
The resulting representation provides complementary views of the same video, ranging from fine-grained clip-level evidence for precise temporal grounding to day-level scaffolds that capture the overall arc of a participant's routine. This multi-scale representation serves as the primary input for all subsequent question generation stages.

\subsection{Stage 2: Query Generation}
\label{app:stage2}
From the structured evidence, we generate candidate multiple-choice questions for each of the three memory types, each associated with a query time at which only prior observations are accessible. 
We use GPT-5.2~\citep{openai2025gpt5}, a different version within the same model family as the captioning stage to avoid self-reinforcing biases. For each memory type, we apply a task-specific pipeline comprising three steps.
\emph{Statement extraction} identifies and aggregates relevant factual statements from the structured evidence to serve as the basis for question formulation. We guide the extractor with memory-type-specific in-context examples: entity-memory prompts demonstrate statements about object states and counts, event-memory prompts demonstrate temporally anchored activities, and behavior-memory prompts demonstrate recurring patterns and cross-event transitions.
\emph{Query generation} formulates a multiple-choice question targeting the relevant memory capability from each extracted statement, together with the ground-truth answer.
\emph{Distractor generation} produces competitive incorrect options for each question. Rather than sampling distractors from unrelated content, we condition the generator on other visual information from the same participant's recording, specifically statements that share salient attributes with the ground truth (such as the same object category, activity, location, or a nearby temporal window) but diverge in the queried dimension. This keeps distractors plausible against the participant's actual memory trace and rules out shortcuts based on global implausibility.

\subsection{Stage 3: Automatic Filtering}
\label{app:stage3}
Raw candidates pass through several filtering stages to ensure that every retained question is genuinely challenging, visually grounded, and non-redundant.

\paragraph{Blind-test filtering.}
We first reject samples that can be answered without video evidence.
A text-only leakage test attempts to answer each question using only the query and options, flagging samples where the correct answer is recoverable from surface cues. We target five such cues: explicit answer leakage in the query text; temporal markers (e.g., ``Day~3'' or ``10:30~AM'') that trivially disambiguate options; lexical shortcut patterns; high token overlap between the query and the correct option relative to the distractors; and common-sense priors that make one option obviously dominant. For episodic memory questions, we additionally apply a stricter model-based variant: an LLM receives only the question and the options under multiple permutations of the option order, and samples whose accuracy across permutations exceeds a fixed threshold are rejected.

\paragraph{Grounded evidence verification.}
Every retained question is verified against the original video to ensure correctness.
We first confirm that each referenced evidence clip exists in the recording and contains enough sampled frames for meaningful visual inspection. 
We then check that the captions associated with each evidence clip are consistent with the question's expected answer, so that the claimed visual evidence actually supports the correct option.
All evidence clips must fall strictly before the question's query timestamp, preserving the constraint that only past observations are accessible. For event-ordering questions, we additionally verify that the temporal order of the referenced clips matches the ground-truth sequence.

\subsection{Stage 4: Human Verification}
\label{app:stage4}

\begin{figure}[t]
  \centering
  \includegraphics[width=\linewidth]{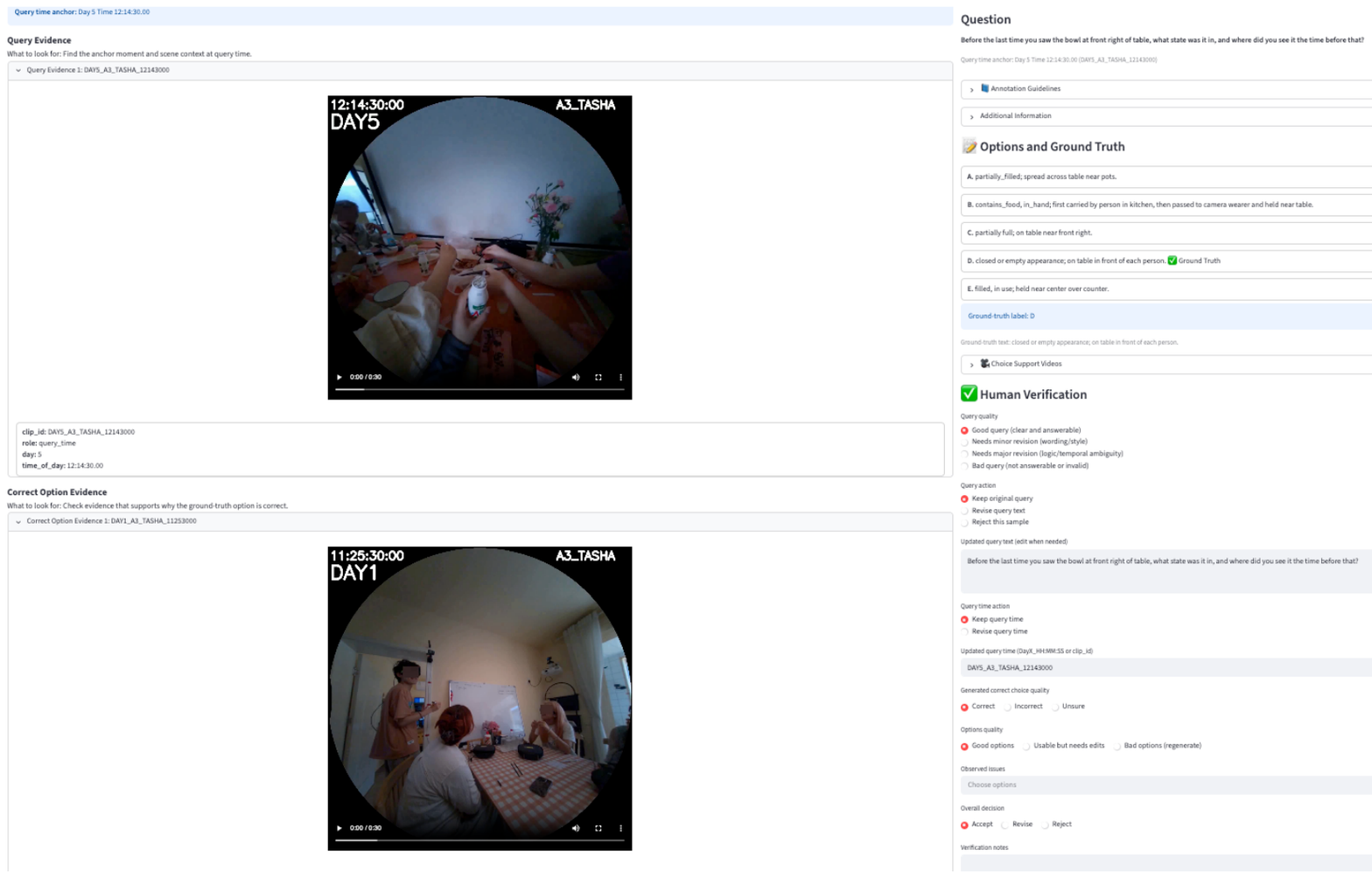}
  \caption{Human verification interface. \textbf{Left:} the annotator reviews the query-time context video and expandable evidence clips alongside the question and option set. \textbf{Right:} multi-dimensional quality assessment
  panel with structured ratings for query quality, correct-choice quality,
  option quality, and an overall accept/revise/reject decision.}
  \label{fig:annotation-ui}
\end{figure}

After automatic filtering, all surviving candidates enter a human verification stage through a purpose-built annotation interface (Figure~\ref{fig:annotation-ui}).

\paragraph{Annotation protocol.}
For each candidate question, the annotator is presented with: (i)~the query-time context video, showing what the camera wearer observes at the moment the question is posed; (ii)~one or more expandable evidence clips drawn from earlier in the timeline, which the pipeline identified as supporting the correct answer; and (iii)~the full set of multiple-choice options with the machine-generated ground-truth label highlighted.
A browsable list of related clips is also available so that annotators can independently verify evidence beyond the pre-selected set.
Annotators follow five guidelines: watch the query context video and at least one relevant support clip; select the best answer based on visual evidence rather than metadata; rate query quality and revise unclear or awkward phrasing; flag option-set issues such as weak, duplicated, or ambiguous distractors; and record an overall decision indicating whether the question can be kept as-is.

\paragraph{Multi-dimensional quality assessment.}
Rather than a single accept or reject label, the interface collects fine-grained judgments along four axes.
\emph{Query quality} is rated on a four-point scale: good (clear and answerable), needs minor revision (wording or style), needs major revision (logic or temporal ambiguity), or bad (not answerable or invalid).
\emph{Correct-choice quality} is assessed as correct, incorrect, or unsure, allowing annotators to flag cases where the machine-generated ground truth does not match the visual evidence.
\emph{Option quality} is rated as good, usable but needs edits, or bad (regenerate), capturing distractor-level issues that may not affect the correct answer itself.
An \emph{observed issues} field records specific failure modes such as temporal inconsistency, duplicate options, or answer leakage.

\paragraph{Revision and decision.}
Based on these assessments, annotators select a query action: keep the original text, revise it with an editable text field for in-place rewording, or reject the sample entirely.
A final overall decision of \textbf{Accept}, \textbf{Revise}, or \textbf{Reject} is recorded alongside free-form verification notes for borderline cases.
For revised samples, the annotator-edited query replaces the machine-generated text, and the updated question undergoes a second round of automatic verification (blind-test and evidence checks) before inclusion in the final benchmark.

\paragraph{Quality loop.}
The full annotation cycle operates as a closed loop: (1)~generate the candidate set, (2)~run strict automatic filtering, (3)~human annotate with multi-dimensional assessment, (4)~apply targeted rewrites on revised samples, and (5)~re-verify and freeze the final release.
We report human acceptance, revision, and rejection rates as part of the benchmark's quality diagnostics.

\paragraph{Tier-2 expert audit.}
Candidates that pass the first-tier annotation undergo a second-tier audit conducted by the authors. Unlike the first tier, which is scoped to the query-time context and the pipeline-selected evidence clips, the second-tier auditor reviews the participant's full multi-day recording before judging each question. This wider context lets the auditor assess the question against the participant's complete activity history rather than a pre-filtered slice, and surface failure modes that are invisible at the clip level: alternative valid answers supported by evidence the pipeline did not flag, distractors that are in fact true of the participant at some other point in the recording, temporal-constraint violations in which a question's correct answer becomes determinable only after the query timestamp, and ambiguities that arise when the same object, location, or activity recurs across days. For each question, the auditor independently re-answers without consulting the proposed ground-truth label, then compares the two; mismatches trigger adjudication and either revision or rejection. The auditor also re-examines the option set for redundancy with the broader recording and rewrites distractors that prove non-competitive once the full context is known. Only questions cleared at this tier are admitted to the final release.

\subsection{Benchmark Details and Statistics}
\label{app: Benchmark Details and Statistics}
We detail the models and key configurations used in each stage of the construction pipeline. In Stage~1 (Evidence Preparation), we use GPT-5~\citep{openai2025gpt5} to generate dense clip-level captions and hierarchical event summaries from the raw egocentric video. In Stage~2 (Query Generation), we use GPT-5.2~\citep{openai2025gpt5} to produce candidate multiple-choice questions for each memory type, conditioned on the structured evidence from Stage~1. In Stage~3 (Automatic Filtering), we employ three models, Gemini-3.1-Pro~\citep{gemini3flash2025}, GPT-5.2~\citep{openai2025gpt5}, and Qwen-3-VL-32B~\citep{Qwen3-VL}, to perform two rounds of filtering: first, a text-only leakage test that removes questions where a majority of the three models can answer correctly without visual input, and second, a quality check that verifies answer correctness, distractor plausibility, and multi-timestamp grounding. In Stage~4 (Human Verification), six annotators at the college or graduate level review each surviving question, spending approximately 20 minutes per sample to assess question clarity, answer correctness, and option quality. Only 15\% of candidates pass this final stage, reflecting the stringent quality standards applied throughout the pipeline.

\section{Additional Implementation Details}
\label{more implementationd details}
Following existing works \citep{yang2025egolife}, in \benchmark{}, each question is associated with a designated query timestamp $t_q$. 
Specifically, only video content observed before $t_q$ is made accessible to the model, ensuring that no future information can be used to answer the question.
Since the videos span multiple days, $t_q$ can range from Day~1 to Day~7, and the temporal gap between the earliest required evidence and $t_q$ routinely exceeds one full day. This constraint is enforced consistently across all evaluated systems.
For all MLLM-based methods, we run experiments for the highest possible frame settings.
For Qwen-3-VL-8B\citep{Qwen3-VL} and Molmo2-8B\citep{molmo2openweightsdata} which we conduct analysis on, we report the best overall performance setting in the main table. 
For agentic approaches, we follow the WorldMM\citep{yeo2025worldmm} to run the caption model with 1FPS sampling on 30s clips.

\section{Additional Experimental Results}
\label{supp:more_experimental_results}

\subsection{Detailed Error Analysis}

\noindent\textbf{Quantitative Error Analysis.}
To better understand the limitations of current MLLMs, we manually inspected 100 benchmark examples and analyzed the common failure patterns. The analysis is based on Gemini-3-Flash responses. Inspired by Video-Holmes \citep{cheng2025video}, we group the failure cases into the following four categories:
\begin{enumerate}[nosep,leftmargin=1.5em]
  \item \emph{Visual Perception Error (VPE)} happens when the model looks at the video but extracts wrong visual details. The necessary visual information for answering the given question is captured, but the model hallucinates on the visual information.
  \item \emph{Visual Omission Error (VOE)} happens when the model simply misses important visual information. This usually happens because the questions require long-range reasoning over the entire video but the current models are restricted by the limited context window.
  \item \emph{Reasoning Error (RE)}) happens when the model sees the right visual details but then makes logical mistakes when processing the visual information. For example, the model correctly tracks the location of the target object across multiple timestamps but synthesizes the temporal information incorrectly.
  \item \emph{Think-Right-Answer-Wrong Error (TRAW)}) happens when the reasoning process is correct and aligns with human reasoning for answering the question, but the model still produces an incorrect prediction.
\end{enumerate}

We find that reasoning Error and Visual Omission Error are the two dominant failure modes, accounting for 32\% of errors, followed by Visual Perception Error at 28\%. 
The relatively small proportion of Think-Right-Answer-Wrong errors (8\%) suggests that when models reason correctly, they usually arrive at the correct answer.
The near-equal split among the three major error types indicates that improving long-horizon video understanding requires advances on multiple fronts: more faithful visual perception, broader temporal coverage to reduce evidence omission, and stronger temporal reasoning capabilities to correctly synthesize information across distant observations.

\subsection{Robustness Analysis}
\label{app:robustness}

\paragraph{Robustness to Filtering Model.}
To assess sensitivity to the choice of filtering model, we randomly sample
100 final benchmark items and rerun the blind filtering stage on their
corresponding pre-human-verification candidates, replacing Qwen-3-VL-32B
with InternVL3.5-38B while keeping GPT-5 and Gemini-3.1-Pro fixed.
Under the alternative filtering ensemble, 93 of the 100 candidates are
retained. On the corresponding final benchmark items, Qwen-3-VL-32B and
Gemini-3-Flash achieve 37.6\% and 39.8\% accuracy, respectively, compared
with 36.8\% and 39.6\% on the full benchmark. These results indicate that
the overall performance trend is stable under the alternative filtering
configuration.

\paragraph{Cross-Model Consistency.}
To examine whether the effect of temporal span is specific to the 8B-scale
model, we further evaluate Qwen-3-VL-32B and Gemini-3-Flash across the same
temporal-certification ranges. As shown in
Table~\ref{tab:cross_model_temporal}, all three models exhibit a consistent
degradation as the required temporal span increases, with the lowest
performance on the longest-range questions. This suggests that the effect of
temporal span persists across different model scales and families.

\begin{table}[h]
\centering
\small
\setlength{\tabcolsep}{7pt}
\begin{tabular}{lccccc}
\toprule
Model & $<$8h & 8--16h & 16--32h & 32h+ & Overall \\
\midrule
Qwen-3-VL-8B
    & 40.3 & 33.7 & 32.5 & 23.2 & 29.6 \\
Qwen-3-VL-32B
    & 46.1 & 40.5 & 38.8 & 31.0 & 36.8 \\
Gemini-3-Flash
    & 48.1 & 46.0 & 41.0 & 32.7 & 39.6 \\
\bottomrule
\end{tabular}
\caption{
Cross-model performance across temporal-certification ranges.
Performance generally decreases as the temporal span required to locate
all supporting evidence increases.
}
\label{tab:cross_model_temporal}
\end{table}

\paragraph{Robustness to Input Scaling.}
We further examine the effect of input frame budget across models of different
scales. As shown in Table~\ref{tab:cross_model_frames}, increasing the number
of input frames generally improves performance for Qwen-3-VL-8B,
Qwen-3-VL-32B, and Gemini-3-Flash, but the gains diminish at larger budgets
and can slightly decrease with further scaling. This consistent pattern
suggests that simply increasing visual input coverage does not fully resolve
the challenges of long-horizon video understanding.

\begin{table}[h]
\centering
\small
\setlength{\tabcolsep}{6pt}
\begin{tabular}{lcccccc}
\toprule
Model & 32 & 64 & 128 & 256 & 512 & 1024 \\
\midrule
Qwen-3-VL-8B
    & 26.0 & 27.4 & 28.5 & 29.6 & 28.4 & 28.3 \\
Qwen-3-VL-32B
    & 30.4 & 32.0 & 35.2 & 35.6 & 36.8 & 36.2 \\
Gemini-3-Flash
    & 32.6 & 33.4 & 36.2 & 39.0 & 39.6 & 37.8 \\
\bottomrule
\end{tabular}
\caption{
Cross-model performance (\%) under different input frame budgets.
Increasing the number of frames generally improves performance up to an
intermediate or high budget, after which the gains plateau or slightly
decrease.
}
\label{tab:cross_model_frames}
\end{table}

\subsection{Additional Benchmark Analysis}
\label{app:oracle}

\begin{figure}[!h]
    \centering

    % ----- Table 2 -----
    \begin{minipage}[t]{0.54\linewidth}
        \vspace{20pt}
        \centering
        \scriptsize
        \resizebox{\linewidth}{!}{
        \begin{tabular}{cc|cccc}
        \toprule
        Trans. & Caption & Entity & Event & Behavior & All \\
        \midrule
        \xmark & \xmark & \textbf{31.5} & 22.0 & 41.0 & 29.6 \\
        \cmark & \xmark & 29.0 & \textbf{23.0} & \textbf{46.0} & \textbf{30.0} \\
        \xmark & \cmark & 29.5 & 21.0 & 45.0 & 29.2 \\
        \cmark & \cmark & \textbf{31.5} & 19.0 & 45.0 & 29.2 \\
        \bottomrule
        \end{tabular}
        }
        \captionof{table}{Effect of auxiliary text inputs (transcript,
        captions) on accuracy (\%).}
        \label{tab:auxiliary}
    \end{minipage}
    \hfill
    % ----- Figure 2 -----
    \begin{minipage}[t]{0.4\linewidth}
        \vspace{0pt}
        \centering
        \includegraphics[width=\linewidth]{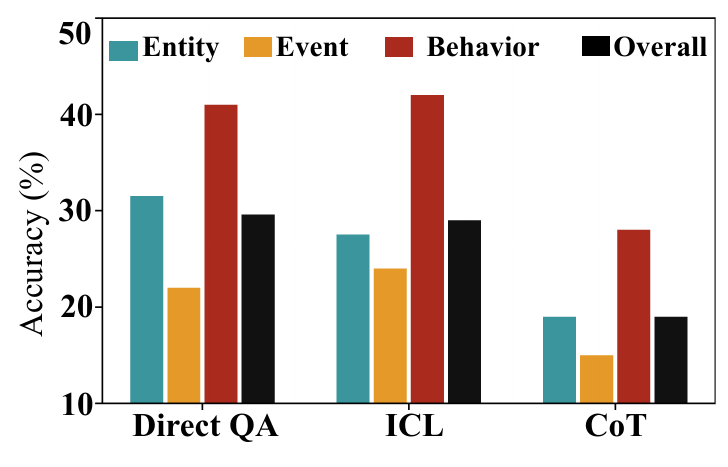}
        \captionof{figure}{Effect of different prompt strategies, including
        direct QA, CoT prompting, and in-context learning.}
        \label{fig:prompting_strategy}
    \end{minipage}

    \vspace{-10pt}
\end{figure}

\noindent\textbf{Auxiliary input information.} 
We study the impact of additional textual inputs (transcripts and captions).
As shown in Table~\ref{tab:auxiliary}, auxiliary text affects each memory type differently. 
Entity Memory is largely insensitive, as precise object states must be read from frames; Event Memory is consistently hurt by captions, likely because dense per-clip captions fragment the cross-clip temporal continuity needed for ordering; and Behavior Memory is the only type that benefits, with transcripts yielding the largest gain, as speech-based signals provide complementary routine and social context.
Despite these per-type differences, the overall gains are marginal: only transcripts yield a small improvement (0.4\%), while captions provide no benefit.
This suggests that additional textual signals alone are insufficient to address the challenges of long-horizon memory reasoning.

\paragraph{Prompting strategy.}
In \Cref{fig:prompting_strategy}, we compare direct QA, in-context learning (ICL), and chain-of-thought (CoT) prompting.
The three strategies expose where the difficulty of long-horizon memory tasks actually lies.
CoT prompting degrades performance substantially across all memory types, cutting overall accuracy by roughly a third and reducing every per-type score, indicating that explicit step-by-step reasoning does not help on memory-intensive tasks and instead amplifies errors that compound over long temporal contexts.
ICL yields performance comparable to direct QA overall, with a small gain on Event memory offset by a modest drop on Entity memory, suggesting that the task format is already well-specified through instructions alone and that additional in-context examples offer little leverage when the underlying challenge is recall rather than format.
Overall, direct QA remains near-optimal, suggesting that more elaborate prompting alone is insufficient to address the dominant challenges in long-horizon video understanding.
This suggests that future improvements may benefit more from enhancing how models encode and access long-horizon visual information than from more sophisticated prompting techniques.

\paragraph{Oracle-Evidence Analysis.}
To disentangle evidence localization from reasoning over the retrieved evidence, we additionally evaluate Gemini-3-Flash by providing only the ground-truth evidence clips required for each question. Its accuracy increases from 39.6\% under the full-video setting to 67.2\% with oracle evidence.
This substantial improvement indicates that locating the relevant evidence
within week-long video is a major source of difficulty, while the remaining gap shows that reasoning over the retrieved multi-event evidence remains challenging.

\paragraph{Human Verification Outcomes.}
Human review substantially modifies the automatically generated candidates: 41.8\% are directly rejected, 6.2\% are refined for wording, 29.4\% undergo substantial revisions to both the question and answer, 11.6\% require updates to the correct answer or option set, and only 11.0\% are retained unchanged.
Among candidates retained after human verification, 70.4\% have their
machine-generated answer supervision substantially revised. These statistics indicate that the final benchmark is extensively human-validated rather than directly inheriting model-generated labels.

\paragraph{Query-Only Evaluation.}
To assess whether the benchmark can be solved through linguistic priors alone, we evaluate Qwen-3-VL-32B and Gemini-3-Flash using only the question and answer options, without any video input. They achieve 18.8\% and 20.4\% accuracy, respectively, close to the 16.8\% random baseline and substantially below their full-video performance of 36.8\% and 39.6\%. This suggests that visual evidence is essential for solving the benchmark.

\paragraph{Agentic Failure Analysis.}
To understand why agentic video frameworks remain limited on
\benchmark{}, we manually analyze 100 failed SiLVR examples by inspecting its retrieved textual memory and reasoning traces. Among these failures, 35\% are caused by retrieval misses, 30\% by perception errors, 32\% by reasoning errors, and 3\% by answer mismatches. These results show that agentic systems fail at multiple stages, suggesting that structured external memory alone does not eliminate errors in evidence access, visual grounding, and downstream temporal reasoning.

\paragraph{Caption Quality.}
We additionally compare our dense captions with the original EgoLife
annotations on 100 randomly sampled clips. Our captions contain 190.7 words, 10.2 distinct objects/entities, and 15.7 action/event verbs on average, compared with 10.8, 1.4, and 1.2, respectively, in the original annotations. This reflects the substantially finer object-, action-, and state-level detail used for constructing \benchmark{}. We will release the intermediate captions, hierarchical summaries, and associated evidence annotations together with the benchmark.

\section{Limitation}
\label{limitations}
Our benchmark is constructed from the week-long recordings in EgoLife, which contain six participants and therefore cover a relatively limited range of individuals, environments, and Behavior routines. This may particularly limit the diversity of Behavior Memory questions. Our goal is to evaluate long-horizon evidence accumulation and reasoning rather than population-level Behavior generalization; extending \benchmark{} to broader participants and environments is an important direction for future work.
Also, while we carefully design tasks to require multi-hop reasoning over temporally distributed evidence, the benchmark is still constructed through controlled generation and filtering procedures. As such, it may not cover the full diversity of real-world long-horizon scenarios. Finally, our evaluation relies on sampled frames and optional auxiliary inputs (e.g., transcripts and captions), which may not fully capture all temporal dynamics in raw video streams. Future work could explore richer video representations and streaming-based evaluation settings.

\end{document}